\documentclass[twocolumn,10pt]{asme2ej}

\usepackage{graphicx} 
\usepackage{hyperref}   
\hypersetup{
	colorlinks=true,
	linkcolor=blue,
	citecolor=blue,
	urlcolor=blue,
}

\usepackage{subcaption}
\usepackage{bm}
\usepackage{amsmath}
\usepackage{amssymb}
\usepackage{cite}

\title{Modeling and Generative-AI-Based Design of Load-Adaptive Gravity Balancing Mechanisms}

\author{Ryotaro Kayawake\thanks{Corresponding author.} \\
    \affiliation{
	PhD Student\\
    Embodied Intelligence and Robotics Laboratory\\
	Graduate School of Information Sciences\\
	Tohoku University\\
	6-6-01, Aramaki, Aoba-ku, Sendai, \\
    Miyagi 9808579, Japan\\
    Email: kayawake.ryotaro.t3@dc.tohoku.ac.jp
    }	
}

\author{Kazuki Abe\\
    \affiliation{
    Assistant Professor\\
	Robot Mechanisms Laboratory\\
	Graduate School of Engineering Science\\
	The University of Osaka\\
	1-3, Machikaneyama, Toyonaka, \\
    Osaka 5608531, Japan\\
    Email: kazuki.abe.org@gmail.com
    }
}

\author{Shota Miyake\\
    \affiliation{
    Specially Appointed Associate Professor (Lecturer)\\
	Robot Mechanisms Laboratory\\
	Graduate School of Engineering Science\\
	The University of Osaka\\
	1-3, Machikaneyama, Toyonaka, \\
    Osaka 5608531, Japan\\
    Email: shota.miyake@pumech.sys.es.osaka-u.ac.jp
    }
}

\author{Masahiro Watanabe\\
    \affiliation{
    Associate Professor\\
	Robot Mechanisms Laboratory\\
	Graduate School of Engineering Science\\
	The University of Osaka\\
	1-3, Machikaneyama, Toyonaka, \\
    Osaka 5608531, Japan\\
    Email: watanabe.masahiro.es@osaka-u.ac.jp
    }
}

\author{Kenjiro Tadakuma\\
    \affiliation{
    Professor\\
	Robot Mechanisms Laboratory\\
	Graduate School of Engineering Science\\
	The University of Osaka\\
	1-3, Machikaneyama, Toyonaka, \\
    Osaka 5608531, Japan\\
    Email: kenjiro.tadakuma.es@osaka-u.ac.jp
    }
}

\begin{document}

\maketitle    

\begin{abstract}
{
Load-adaptive gravity balancing mechanisms (LA-GBMs) can accommodate various loading conditions by passively changing their characteristics in response to payload variations. 
However, their design is difficult because both the desired mechanism motion and static equilibrium under variable payloads must be satisfied simultaneously.
This study proposes a general design methodology for LA-GBMs that does not depend on specific mechanism architectures or mechanical elements. 
The necessary conditions for the potential fields of LA-GBMs are formulated, and two general forms are derived: an affine form representing the effect of payload mass and a factorized form representing state transitions associated with load adaptation and gravity balancing.
These forms are then provided to generative AI as design requirements to generate candidate potential functions. 
The generated functions are analytically verified in terms of their conformity to the two general forms and the conditions required for valid LA-GBMs. 
Furthermore, the obtained potential functions are decomposed into individual terms, and an example of a method for constructing an LA-GBM by combining springs, counterweights, and function-generating linkage mechanisms is presented.
By using potential functions as an intermediate representation, the proposed framework enables the generation of LA-GBM design candidates without prescribing a mechanism architecture in advance. Mechanical realizability and manufacturability of the generated potential fields remain important issues for future work.
}
\end{abstract}

\begin{nomenclature}
\entry{$g$}{gravitational acceleration.}
\entry{$k$}{spring stiffness.}
\entry{$l$}{the length of the lever arm and the mounting height of the spring.}
\entry{$M$}{payload mass.}
\entry{$\bm{p}$}{passive variables of the system.}
\entry{$p$}{slider displacement.}
\entry{$\bm{q}$}{active variables of the system.}
\entry{$q$}{lever-arm angle.}
\entry{$\bm{t}$}{ alternative representation of the passive variable $\bm{p}$.}
\entry{$U$}{ smooth potential field formed by an
LA-GBM.}
\entry{$U_g$}{gravitational potential energy of the payload.}
\entry{$U_c$}{compensating potential field.}
\entry{$\Theta$}{the range of payload masses.}
\entry{$\pi_x(\Omega)$}{the projection of the feasible domain $\Omega$ onto the $x$.}
\entry{$\Omega$}{subset that represents constraints on the Cartesian product space of the active variables $\bm{q}$ and passive variables $\bm{p}$.}
\entry{$\Omega_x$}{the fiber of $\Omega$ over $\bm{q}$ with respect to $\bm{p}$.}
\end{nomenclature}

\section{Introduction}
Gravity balancing mechanisms (GBMs) are effective for systems operating under gravity because they can reduce actuator loads, improve energy efficiency, and enhance safety in human--robot interaction \cite{review_0}. 
Various approaches have been proposed for the implementation of GBMs \cite{review_1,review_2,review_3,review_4,review_5,review_6}, among which counterweights \cite{histry_1,histry_2,histry_3}, springs \cite{histry_4,histry_5,histry_6,histry_7,histry_8,histry_9}, and buoyancy \cite{histry_10} are commonly employed.

In recent years, GBMs capable of adapting to changes in payload have been actively investigated for applications involving the handling of various objects \cite{review_0}.
Such mechanisms are referred to this study as load-adaptive gravity balancing mechanisms (LA-GBMs), and several approaches have been reported. 
Examples include methods that vary the preload of springs \cite{SRGCM_yoatsu_1,SRGCM_yoatsu_2,SRGCM_yoatsu_3}, vary the transmission ratio according to the applied load \cite{SRGCM_gensoku_1,SRGCM_gensoku_2,SRGCM_gensoku_3}, and vary the point of force application \cite{SRGCM_1,SRGCM_2,SRGCM_3,SRGCM_4,SRGCM_5,SRGCM_6,SRGCM_7,SRGCM_8}.

However, one of the major challenges in the development of GBMs is the complexity of mechanism design.
To realize gravity balancing, a mechanism must not only satisfy the kinematic conditions required to generate the desired motion but also appropriately generate balancing forces or moments that compensate for gravity at each configuration. 
In particular, when elastic elements such as springs are used for gravity compensation, parameters affecting the static equilibrium conditions, such as the attachment positions and deformations of the elastic elements, are determined by the kinematics of the mechanism. 
Consequently, these design variables generally cannot be determined independently \cite{review_0}, and the design of a GBM must therefore be treated as an integrated problem in which kinematic and static conditions are mutually coupled.

To address this difficulty, several generalized approaches to GBM design have been investigated. 
For example, methods based on potential energy or stiffness matrices have been proposed to systematically determine design parameters such as spring arrangements and link lengths \cite{Design_1,Design_2,Design_3}.
These methods form a basis for analytical GBM design by deriving appropriate arrangements of mechanical elements for a prescribed linkage from the structure of the potential energy of the system.
Other approaches generate the joint arrangement of a mechanism suitable for gravity balancing itself through optimization \cite{Design_4,Design_5}. 
These methods provide a framework for mechanism synthesis in which candidate joint arrangements are generated and compared under predefined components and design rules, followed by optimization of their design parameters.
For LA-GBMs, a design method has also been proposed in which design parameters dependent on payload mass are extracted and the arrangements of load-adaptive elements and balancing springs are optimized \cite{Design_6}.

However, most existing GBM design theories prescribe the fundamental joint arrangement, mechanical architecture, or available mechanical elements in advance.
Consequently, the resulting design solutions are restricted to the range of predefined structures or candidate mechanisms. 
This limitation is particularly pronounced for LA-GBMs, which are subject to a larger number of functional constraints.

Accordingly, this study aims to establish a design methodology for generating novel LA-GBM architectures by expressing the mechanical characteristics required of LA-GBMs using potential functions generalized in a new form and employing these functions as intermediate representations. 
Artificial intelligence (AI) is incorporated as part of the design process to assist in exploring design possibilities based on these potential functions.
This approach may enable the design of novel GBMs that would be difficult to obtain through conventional searches relying primarily on designers' experience or extensions of existing mechanisms.

The main contributions of this study are as follows:
\begin{enumerate}
\item The potential field of an LA-GBM is formulated in a new general form that is independent of specific mechanism architectures and constituent elements.
\label{koken_1}
\item A novel AI-based design framework is presented for the structural synthesis of LA-GBMs.
\label{koken_2}
\end{enumerate}

\section{Theory}
This section formulates the potential fields of LA-GBMs in a new general form that is independent of specific mechanism architectures and constituent elements.
Specifically, the potential field of an LA-GBM is generalized as a class of functions that simultaneously satisfies two forms introduced later: the affine form and the factorized form.
First, based on the characteristics of existing LA-GBMs, the class of mechanisms considered in this generalization is clearly defined. The affine and factorized forms are then derived as classes of potential functions that satisfy these definitions.

\subsection{Definitions of Variables and Sets}
This section defines the principal variables and the associated sets used in the analysis and formulation of the conditions.
First, the smooth potential field formed by an LA-GBM is expressed as $U = U(\bm{q}, \bm{p};M)\in \mathbb{R}$.
Here, $\bm{q} \in \mathbb{R}^m$ denotes the $m$ active variables of the system, and $\bm{p} \in \mathbb{R}^n$ denotes the $n$ passive variables of the system. 
The active variables $\bm{q}$ are defined as state variables that can be arbitrarily specified by the user, whereas the passive variables $\bm{p}$ are defined as internal state variables whose states are passively determined by the equilibrium conditions of the potential.
Examples of the active variables $\bm{q}$ include the joint angle of a lever arm and the vertical displacement of a lift, whereas an example of the passive variables $\bm{p}$ is the extension of a spring supporting a payload.
$M\in \mathbb{R}$ denotes the payload mass, and the potential $U$ constitutes a family of functions parameterized by $M$.
In addition, let $U_g$ denote the gravitational potential energy of the payload $M$, and let $U_c$ denote the compensating potential field generated by mechanical elements constituting the LA-GBM, such as counterweights and springs, to compensate for $U_g$.
Similarly, $U_g$ and $U_c$ are expressed as families of functions parameterized by the payload mass $M$, namely, $U_g = U_g(\bm{q},\bm{p};M)\in \mathbb{R}$ and $U_c = U_c(\bm{q},\bm{p};M)\in \mathbb{R}$.

Next, several sets associated with the variables defined above are introduced.
First, the range of the payload mass $M$ is defined as $\Theta \subset \mathbb{R}_{> 0}$, a subset of the positive real numbers.
This set represents the range of payload masses to which the GBM can adapt.
The feasible domain of the state variables $\bm{q} \in \mathbb{R}^m$ and $\bm{p} \in \mathbb{R}^n$ is denoted by $\Omega \subset \mathbb{R}^{m+n}$.
The domain $\Omega$ is a subset that represents constraints on the Cartesian product space of the active variables $\bm{q}$ and passive variables $\bm{p}$, such that $(\bm{q},\bm{p}) \in \Omega$.

\begin{figure}[t]
    \centering
    \includegraphics[width=1.0\linewidth, trim=30 270 30 270, clip]{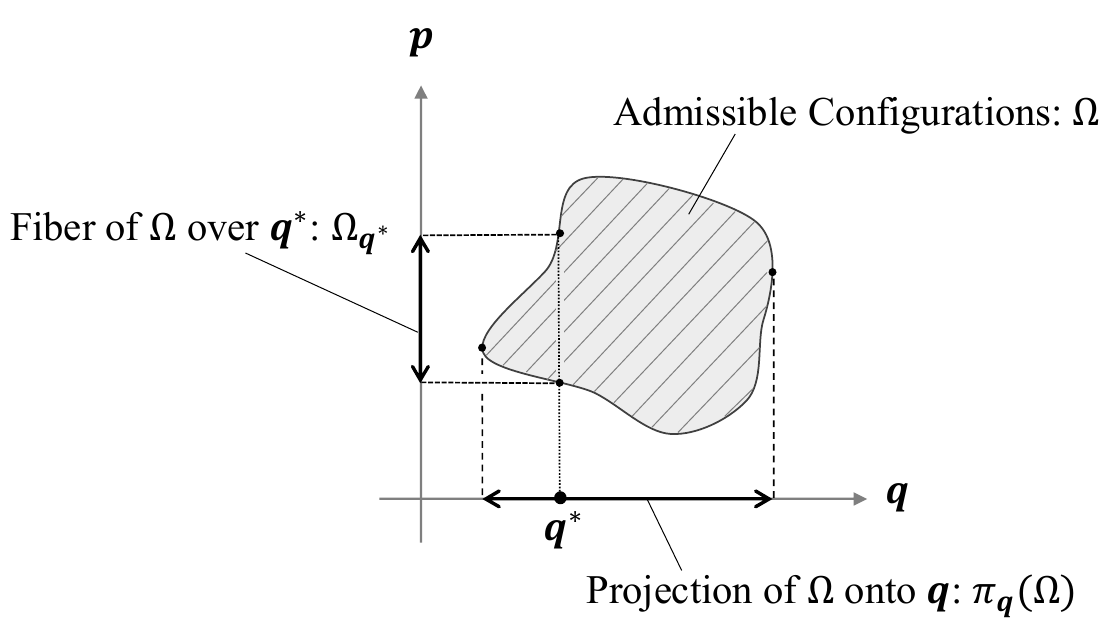}
    \caption{
    Definition of the admissible region $\Omega$, its projection $\pi_{\bm{q}}(\Omega)$, and the fiber $\Omega_{\bm{q}}$.
    }
    \label{fig_def_sets}
\end{figure}

Furthermore, because the active variables $\bm{q}$ and passive variables $\bm{p}$ must be treated independently in the subsequent proof, the corresponding projection and fiber are formulated as follows (Fig. \ref{fig_def_sets}).
First, the projection of the feasible domain $\Omega$ onto the $\bm{q}$-space is defined as
\begin{equation}
\pi_{\bm{q}}(\Omega) 
:=
\{\bm{q} \ | \ \exists \bm{p}, (\bm{q},\bm{p}) \in \Omega \} \subset \mathbb{R}^{m}
\label{Eq_pr_0_def_1}
\end{equation}
This projection corresponds to the set of admissible values of $\bm{q}$ contained in the feasible domain $\Omega$.
In addition, the fiber of $\Omega$ over $\bm{q}$ with respect to $\bm{p}$ is defined as
\begin{equation}
\Omega_{\bm{q}} 
:=
\{\bm{p} \in \mathbb{R}^n \ | \ (\bm{q},\bm{p}) \in \Omega \} \subset \mathbb{R}^{n}
\label{Eq_pr_0_def_2}
\end{equation}
This fiber corresponds to the set of admissible states of the passive variables $\bm{p}$ within $\Omega$ when the active variables $\bm{q}$ are fixed at a given value.

\begin{figure*}[t!]
    \centering

    \begin{subfigure}[b]{0.45\linewidth}
        \centering
        \includegraphics[
            width=\linewidth,
            trim=150 330 180 330,
            clip
        ]{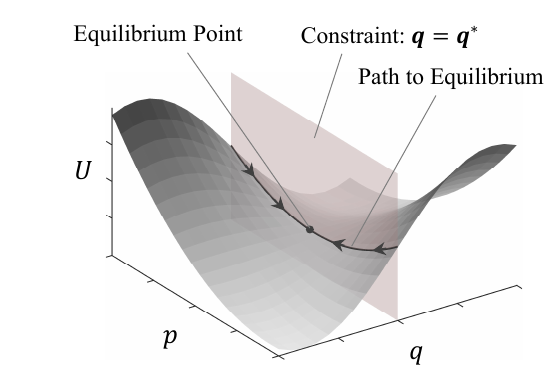}
        \caption{Payload Adaptation.}
        \label{fig_previous_motion_a}
    \end{subfigure}
    \hfill
    \begin{subfigure}[b]{0.45\linewidth}
        \centering
        \includegraphics[
            width=\linewidth,
            trim=180 330 150 330,
            clip
        ]{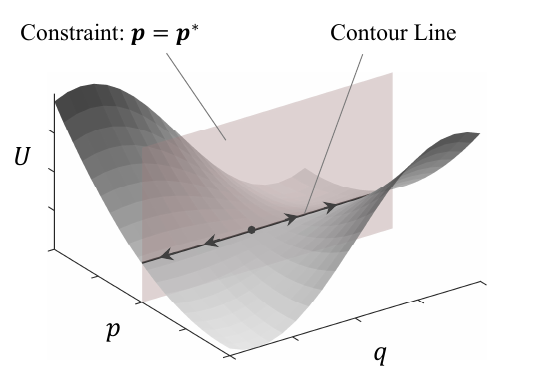}
        \caption{Gravity Balancing.}
        \label{fig_previous_motion_b}
    \end{subfigure}

    \caption{
    Potential field of an existing LA-GBM \cite{SRGCM_5}, illustrating one example that satisfies the condition in Eq. (\ref{Eq_pr_12}). The path toward the local minimum represents the load-adaptation process, whereas the path along which the potential remains at a constant minimum represents the gravity-balancing process.
    }
    \label{fig_previous_motion}
\end{figure*}

\begin{figure}[!t]
    \centering
    \includegraphics[width=0.95\linewidth, trim=100 340 100 340, clip]{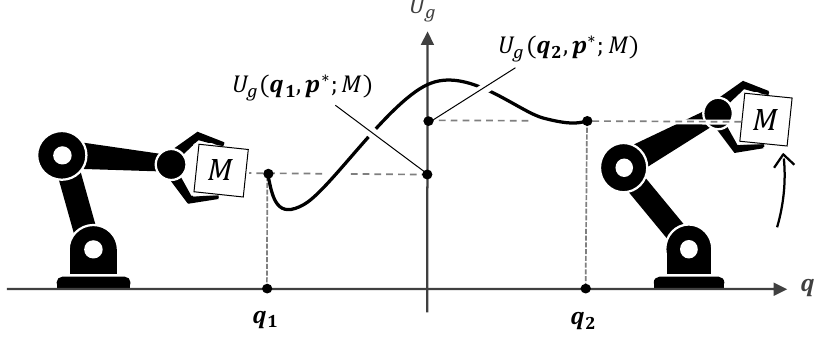}
    \caption{Schematic illustration of a state variation satisfying the condition $dU_g \neq 0$ in Eq. (\ref{Eq_pr_13})}
    \label{fig_mochiage}
\end{figure}

\begin{figure}[t]
    \centering
    \includegraphics[width=0.7\linewidth, trim=170 320 170 300, clip]{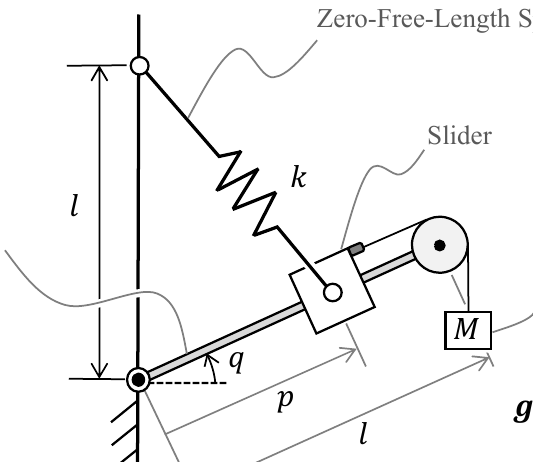}
    \caption{
    Example of an LA-GBM consisting of a lever arm, a slider, and a zero-free-length spring.
A mechanism based on this configuration, incorporating pulleys to adjust the transmission ratio, was proposed in Ref. \cite{SRGCM_5}.
}
    \label{fig_fabricated_1}
\end{figure}

\begin{figure*}[!t]
    \centering
    \begin{subfigure}[b]{0.24\linewidth}
        \centering
        \includegraphics[
            width=\linewidth,
            trim=180 280 180 280,
            clip
        ]{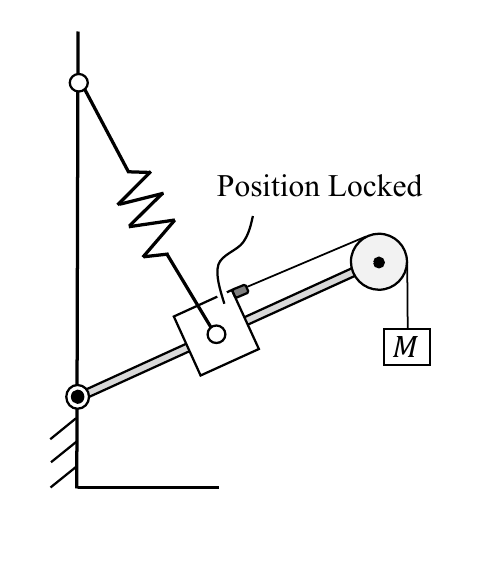}
        \caption{Initial state}
        \label{fig_a}
    \end{subfigure}
    \hfill
    \begin{subfigure}[b]{0.24\linewidth}
        \centering
        \includegraphics[
            width=\linewidth,
            trim=180 280 180 280,
            clip
        ]{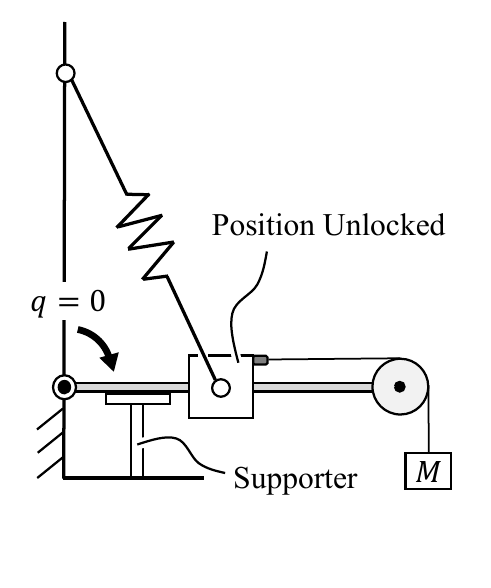}
        \caption{Fixing the arm angle}
        \label{fig_b}
    \end{subfigure}
    \hfill
    \begin{subfigure}[b]{0.24\linewidth}
        \centering
        \includegraphics[
            width=\linewidth,
            trim=180 280 180 280,
            clip
        ]{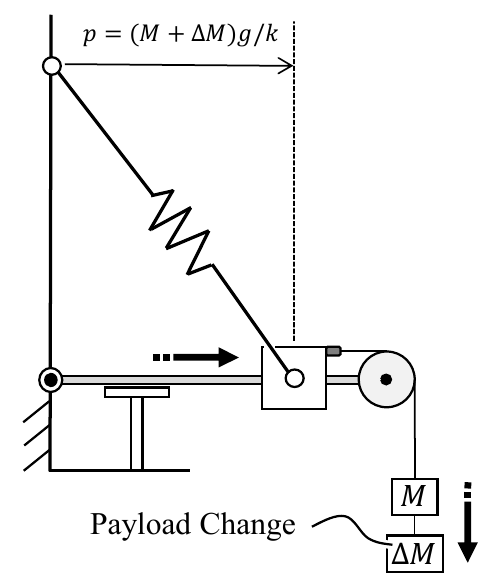}
        \caption{Load-adaptive motion}
        \label{fig_c}
    \end{subfigure}
    \hfill
    \begin{subfigure}[b]{0.24\linewidth}
        \centering
        \includegraphics[
            width=\linewidth,
            trim=180 280 180 280,
            clip
        ]{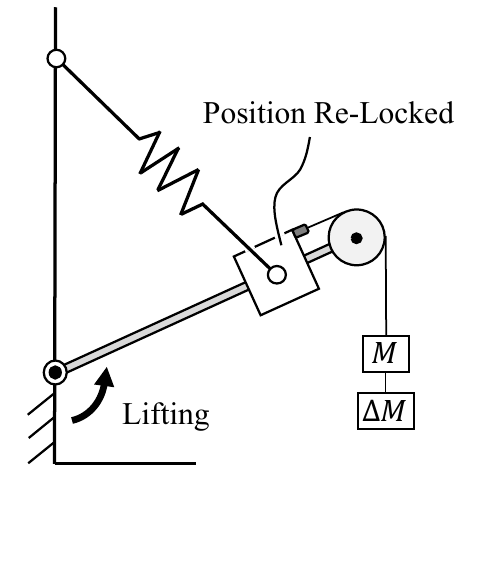}
        \caption{Gravity-balancing motion}
        \label{fig_d}
    \end{subfigure}

    \caption{
    Sequence of operations of an LA-GBM consisting of a lever arm, a slider, and a zero-free-length spring, from load adaptation to gravity balancing.
The initial state is defined as the state in which the slider position $p$ is locked (a). 
First, the lever arm is set to the horizontal configuration $q = 0$ and supported by a base or other support, while the lock on the slider position $p$ is released (b). 
Next, when an additional payload $\Delta M$ is applied, the slider moves to the appropriate equilibrium position $p = (M+\Delta M)g/k$ corresponding to the total payload $M+\Delta M$ (c). 
Finally, by locking the slider again at $p = (M+\Delta M)g/k$, the gravitational potential energy $U_g$ of the payload is compensated by the elastic potential energy $U_c$ stored in the spring for any arm angle $q$, allowing the payload displacement to be varied without energy consumption (d).
    }
    \label{fig_lever_arm_motion}
\end{figure*}

\subsection{Definition and Scope of the Mechanisms Considered}
The generalization of the potential fields developed in this study focuses on LA-GBMs that achieve load adaptation by utilizing locking elements or similar components to appropriately constrain selected variables, thereby allowing internal states, such as spring deformation, to change passively in response to an unknown payload.
This scope is motivated by existing LA-GBMs, which generally require auxiliary operations, such as fixing the rotational angle of a link, during the load-adaptation process \cite{SRGCM_1,SRGCM_2,SRGCM_3,SRGCM_4,SRGCM_5,SRGCM_6,SRGCM_7,SRGCM_8}.
For example, an LA-GBM in which the payload $M$ is varied while a lever arm is fixed in a horizontal orientation and the point of application of the spring force is subsequently determined through load adaptation \cite{SRGCM_5} falls within this scope.
The conditions that the potential fields of such LA-GBMs must satisfy are expressed by the following two mathematical conditions, and the mechanisms considered in this study are restricted to those satisfying both conditions.
\begin{equation}
\begin{aligned}
&\forall M \in \Theta,\ 
\exists \bm{q}^{*} \in \pi_{\bm{q}}(\Omega),\ 
\exists \bm{p}^* \in \Omega_{\bm{q}}, \\
&\Big(
\bm{p}^* \in
\operatorname*{arg\,locmin}_{\bm{p} \in \Omega_{\bm{q}^{*}}}
U(\bm{q}^{*},\bm{p};M)
\Big)\\
&\qquad \wedge
\Big(
\forall \bm{q} \in \pi_{\bm{q}}(\Omega),\
U(\bm{q},\bm{p}^*;M)=U^*
\Big),
\end{aligned}
\label{Eq_pr_12}
\end{equation}
\begin{equation}
\begin{aligned}
&\forall M \in \Theta,\ 
\exists \bm{q}^{*} \in \pi_{\bm{q}}(\Omega),\ 
\exists \bm{p}^* \in \Omega_{\bm{q}}, \\
&\Big(
\bm{p}^* \in
\operatorname*{arg\,locmin}_{\bm{p} \in \Omega_{\bm{q}^{*}}}
U(\bm{q}^{*},\bm{p};M)
\Big)\\
&\qquad \wedge
\Big(
\exists \bm{q} \in \pi_{\bm{q}}(\Omega),\
\nabla_{\bm{q}} U_g(\bm{q},\bm{p}^*;M)
\neq \bm{0}
\Big).
\end{aligned}
\label{Eq_pr_13}
\end{equation}
The term $\operatorname*{arg\,locmin}\limits_{\bm{p} \in \Omega_{\bm{q}^{*}}} U(\bm{q}^{*},\bm{p};M)$ in the equation denotes the set of $\bm{p}^{*}$ at which the potential $U$ attains a local minimum.

The condition in Eq. (\ref{Eq_pr_12}) means that the potential $U$ of the LA-GBM attains a local minimum $U^*$ on a hyperplane where the active variables $\bm{q}$ are fixed at a certain value $\bm{q}^{*}$, and that, once the passive variables $\bm{p}$ are fixed at $\bm{p}=\bm{p}^{*}$, the potential $U$ remains at the same minimum value $U^*$ regardless of the values of the active variables $\bm{q}$.
This condition represents the load-adaptation process along a path toward the local minimum value $U^*$ and the gravity-balancing process along a path on which the potential remains at $U^*$ (Fig. \ref{fig_previous_motion}), and is used in the derivation of the factorized form in Sec. 2.4.
In contrast, the condition in Eq. (\ref{Eq_pr_13}) is introduced as a functional requirement of the LA-GBM and requires that the mechanism be capable of lifting the payload during the gravity-balancing process (Fig. \ref{fig_mochiage}).

As an example of an existing mechanism, the lever-arm mechanism shown in Fig. \ref{fig_fabricated_1} is examined to confirm that it functions as a defined LA-GBM.
In this mechanism, the slider is directly connected to the payload $M$, and the payload height is given by the height of the tip of the lever arm minus the slider displacement.
Accordingly, the potential field generated by this mechanism is expressed as follows.
\begin{equation}
U(q,p;M)
=
Mg(l \sin{q} - p)
+
\frac{1}{2}k{(l^2 + p^2 - 2lp \sin{q})}
\label{Eq_pr_14}
\end{equation}
Here, $g$ denotes the gravitational acceleration, $l$ denotes the length of the lever arm and the mounting height of the spring, and $k$ denotes the spring stiffness. 
In addition, the lever-arm angle $q$ is used as the active variable $\bm{q}$ of the mechanism, while the slider displacement $p$ is used as the passive variable $\bm{p}$. 
The movable domain $\Omega$ is defined as
\begin{equation}
\Omega
=
\left\{
(q,p)\in\mathbb{R}^2
\mid
-\pi/2 < q < \pi/2,\;
0 \leq p \leq l
\right\}
,
\ \ \
Mg/k < l
\label{Eq_pr_15}
\end{equation}
First, under the constraint condition that keeps the lever arm horizontal, $q = 0$, Eq. (\ref{Eq_pr_14}) becomes
\begin{equation}
\frac{\partial U}{\partial p}(0,p;M)
=
k p
- 
M g
\label{Eq_pr_18}
\end{equation}
\begin{equation}
\frac{\partial^2 U}{\partial p^2}(0,p;M)
=
k
>
0
\label{Eq_pr_19}
\end{equation}
This means that, under the constraint condition $q = 0$, the passive variable $p$ moves toward the local minimum point $p = Mg/k$.
Furthermore, when the passive variable $p$ is fixed at the equilibrium point $p = Mg/k$ obtained through the load-adaptive operation,
\begin{equation}
\frac{\partial U}{\partial q}(q,Mg/k;M)
=
(Mgl - klp) \cos{q}
\big |_{p = Mg/k}
=
0
\label{Eq_pr_20}
\end{equation}
holds for any $q$ within the movable domain.
In other words, when $p$ is fixed at $p = Mg/k$, the potential $U$ remains constant regardless of $q$.
From Eqs. (\ref{Eq_pr_18})--(\ref{Eq_pr_20}), the potential $U$ formed by this mechanism satisfies condition (\ref{Eq_pr_12}).

Furthermore, since the potential energy $U_g$ of the payload $M$ satisfies
\begin{equation}
\frac{\partial U_g}{\partial q}(q,Mg/k;M)
=
M
\frac{\partial f}{\partial q}(q,Mg/k)
=
Mgl \cos{q}
\neq 
0
,
\label{Eq_pr_21}
\end{equation}
the potential $U$ formed by this mechanism also satisfies condition (\ref{Eq_pr_13}).

In the following, the affine form and the factorized form, which represent general forms of the potential $U$ of mechanisms such as the example above, are derived in Secs. 2.3 and 2.4, respectively.

\subsection{Affine Formulation}
In this section, the potential field $U$ formed by an LA-GBM is generalized as an affine function with respect to the payload $M$.
The generalization in affine form explicitly shows that the potential $U$ formed by an LA-GBM can be represented as the superposition of the potential energy $U_g$ associated with the payload $M$ and the compensating potential $U_c$ generated by elements such as counterweights and springs.

The affine form of the potential field generated by an LA-GBM is given by the following function.
\begin{equation}
U(\bm{q},\bm{p};M)
=
M
f(\bm{q},\bm{p})
+
h(\bm{q},\bm{p})
\label{Eq_pr_4}
\end{equation}
Here, $f(\bm{q},\bm{p})$ and $h(\bm{q},\bm{p})$ are arbitrary smooth functions of the state variables $\bm{q}$ and $\bm{p}$.
Equation (\ref{Eq_pr_4}) is an extension of the law of conservation of mechanical energy under quasi-static operation, which is known as one of the most fundamental properties satisfied by GBMs \cite{review_0}\cite{SRGCM_1},
\begin{equation}
U
=
U_g + U_c
\label{Eq_pr_1}
\end{equation}
with the additional consideration of the effect of the payload $M$.
The derivation of Eq. (\ref{Eq_pr_4}) is presented below.

First, the potential energy $U_g$ of the payload can be expressed as the product of the payload $M$ and the term $f(\bm{q}, \bm{p})$ corresponding to its vertical displacement, as follows.
\begin{equation}
U_g(\bm{q},\bm{p};M)
=
M
f(\bm{q},\bm{p})
.
\label{Eq_pr_2}
\end{equation}
Furthermore, when the payload $M$ varies from task to task, mechanism parameters such as the spring stiffness and the counterweight mass cannot be adjusted according to the payload $M$, and all resulting changes in the mechanism state are reflected in the state variables $\bm{q}$ and $\bm{p}$.
Therefore, the compensating potential $U_c$ of an LA-GBM without additional operations can be described as follows by a function $h = h(\bm{q}, \bm{p})$ whose arguments consist only of the state variables.
\begin{equation}
U_c(\bm{q},\bm{p};M)
=
h(\bm{q},\bm{p})
.
\label{Eq_pr_3}
\end{equation}
Finally, by substituting Eqs. (\ref{Eq_pr_2}) and (\ref{Eq_pr_3}) into Eq. (\ref{Eq_pr_1}), Eq. (\ref{Eq_pr_4}) is obtained as the general form of the potential field $U$ of an LA-GBM that handles an unknown payload $M$.

Furthermore, under Eq. (\ref{Eq_pr_4}), condition (\ref{Eq_pr_13}) can be expressed in terms of the function $f$ as follows.
\begin{equation}
\exists \bm{q} \in \pi_{\bm{q}}(\Omega)
,
\nabla_{\bm{q}} f(\bm{q}, \bm{p}^*) \neq \bm{0}
.
\label{Eq_pr_5}
\end{equation}

\begin{figure*}[h!]
    \centering
    \includegraphics[width=0.9\linewidth, trim=0 10 0 10, clip]{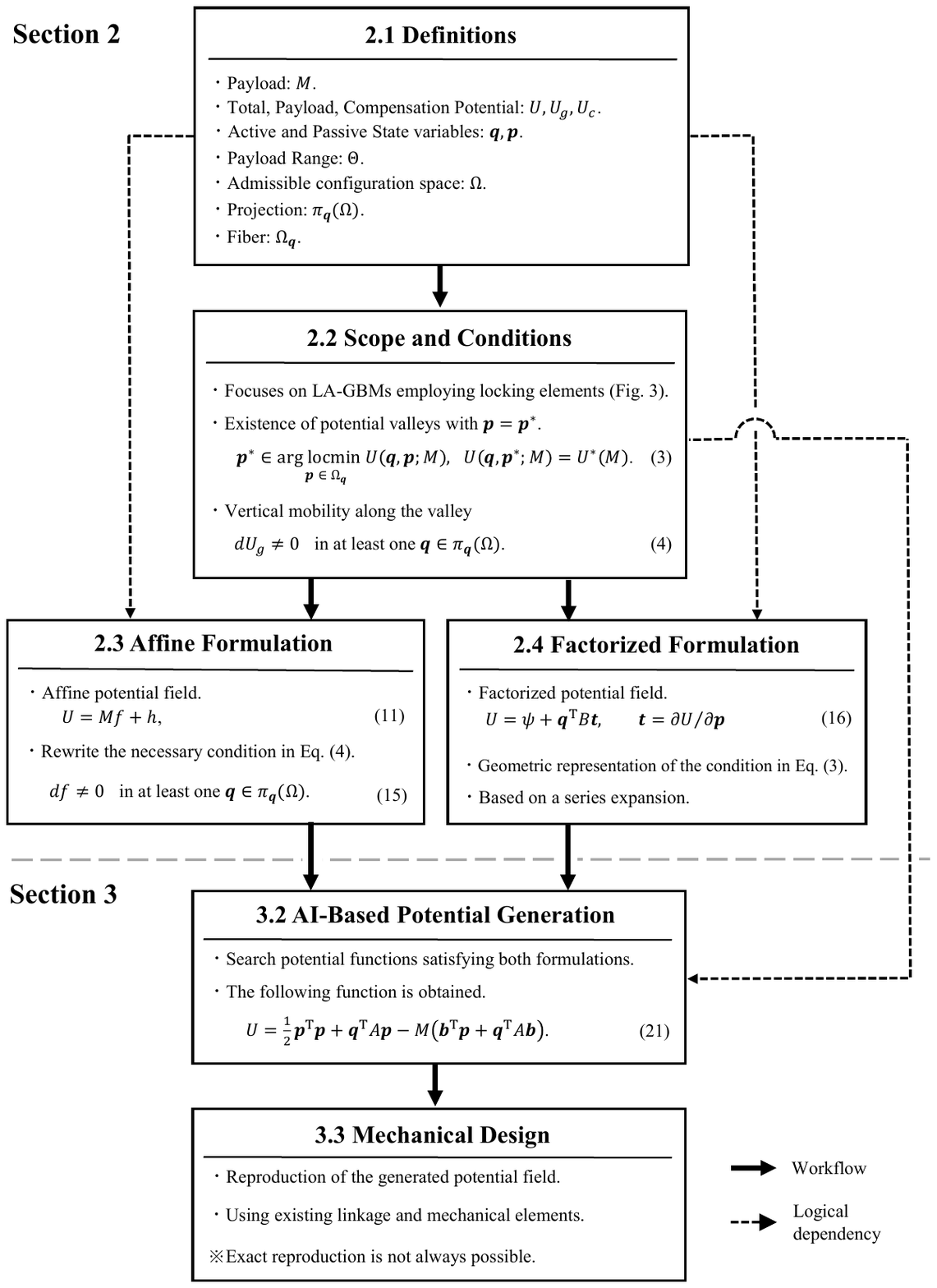}
    \caption{
    Design flow for generating candidate potential functions of LA-GBMs using generative AI based on the affine and factorized forms derived in Sec. 2 and developing them into mechanism structures.
    }
    \label{fig_flow_AI}
\end{figure*}

\subsection{Factorized Formulation}
In this section, the potential field $U$ formed by an LA-GBM is generalized in a factorized form.
The generalization in factorized form explicitly represents the load-adaptive operation as a state transition toward an equilibrium point of the potential field and the gravity-balancing operation as a state transition along a contour of the potential field.

The factorized form is given by the following function.
\begin{equation}
U(\bm{q},\bm{t};M)
=
\psi (\bm{t};M)
+
\bm{q}^\mathrm{T}
B(\bm{q},\bm{t};M)
\bm{t}
.
\label{Eq_new_1}
\end{equation}
Here, $\psi (\bm{t};M) \in \mathbb{R}$ and $B(\bm{q},\bm{t};M) \in \mathbb{R}^{m \times n}$ are arbitrary functions, and the variable $\bm{t}$ is an alternative representation of the passive variable $\bm{p}$ defined by the following variable transformation.
\begin{equation}
\exists \bm{q}^*
,
\bm{t}
=
\frac{\partial U}{\partial \bm{p}}(\bm{q}^*, \bm{p};M)
.
\label{Eq_new_2}
\end{equation}
It is assumed that, in a neighborhood of $\bm{p} = \bm{p}^{*}$ satisfying $\bm{t} = \bm{0}$, the Hessian matrix $\nabla_{\bm{p}}^2 U(\bm{q}^*,\bm{p};M)$ is positive definite and the variable transformation is regular.
The derivation of Eq. (\ref{Eq_new_1}) is presented below.

First, the potential $U$ is expressed as the following series in the form of a polynomial in $\bm{q}$ and $\bm{t}$.
\begin{equation}
U(\bm{q},\bm{t};M)
=
\phi(\bm{q};M)
+
\psi (\bm{t};M)
+
\bm{q}^\mathrm{T}
B(\bm{q},\bm{t};M)
\bm{t}
.
\label{Eq_new_3}
\end{equation}
The load-adaptive operation can be interpreted as a state transition toward the equilibrium point $\partial U/\partial \bm{p} (\bm{q}^, \bm{p}^; M) = \bm{0}$ on the hyperplane where the active variable $\bm{q}$ is fixed at $\bm{q} = \bm{q}^*$.
That is, when there exists a state $(\bm{q}^*, \bm{p}^*)$ satisfying $\bm{t}(\bm{q}^*, \bm{p}^*;M) = \bm{0}$, the potential $U(\bm{q},\bm{t};M)$ attains the local minimum value $U^*(M) = U(\bm{q}^,\bm{0};M)$.
Furthermore, during the gravity-balancing operation, the state must transition along a contour on the hyperplane where the passive variable $\bm{p}$ is fixed at $\bm{p} = \bm{p}^*$. Therefore, the following condition must be satisfied:
\begin{equation}
\forall \bm{q} \in \pi_{\bm{q}}(\Omega) 
, \ \
U(\bm{q},\bm{0};M)
=
U^*(M)
.
\label{Eq_new_4}
\end{equation}

Substituting the condition in Eq. (\ref{Eq_new_4}) into Eq. (\ref{Eq_new_3}) eliminates the third term on the right-hand side, indicating that $\phi(\bm{q};M)$ is a constant function.
\begin{equation}
\begin{aligned}
U(\bm{q},\bm{0};M)
&=
\phi(\bm{q};M)
+
\psi (\bm{0};M)
=
U^*(M) 
\\
\therefore
\phi(\bm{q};M)
&=
\phi(M)
.
\end{aligned}
\label{Eq_new_5}
\end{equation}
In this case, $\phi(M)$ can be absorbed into the function $\psi (\bm{t};M)$ by applying the mapping $\phi(M) + \psi (\bm{q};M) \mapsto \psi (\bm{q};M)$. Consequently, the factorized form in Eq. (\ref{Eq_new_1}) is obtained.

\begin{figure*}[!t]
    \centering

    \begin{subfigure}[b]{0.45\linewidth}
        \centering
        \includegraphics[
            width=\linewidth,
            trim=150 330 180 330,
            clip
        ]{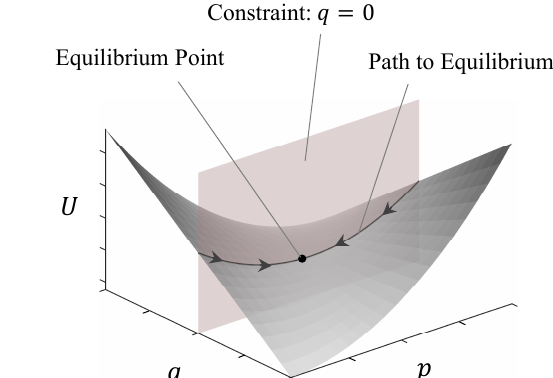}
        \caption{Payload Adaptation.}
        \label{fig_generated_potential_a}
    \end{subfigure}
    \hfill
    \begin{subfigure}[b]{0.45\linewidth}
        \centering
        \includegraphics[
            width=\linewidth,
            trim=180 330 150 330,
            clip
        ]{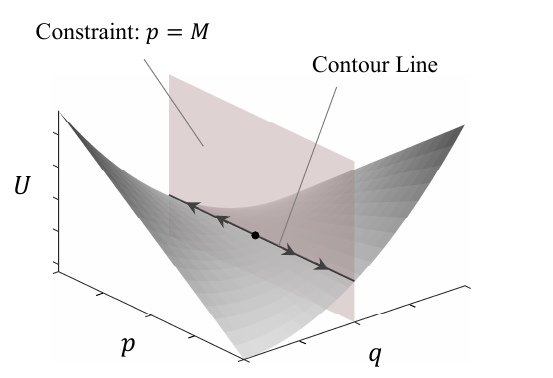}
        \caption{Gravity Balancing.}
        \label{fig_generated_potential_b}
    \end{subfigure}

    \caption{
    Schematic shape of the potential field obtained using generative AI. The surface is plotted using Eq. (\ref{Eq_AI_9}), which corresponds to Eq. (\ref{Eq_AI_1}) for $\bm{q} \in \mathbb{R}$ and $\bm{p} \in \mathbb{R}$.
    }
    \label{fig_generated_motion}
\end{figure*}

\section{Design of LA-GBMs Using Generative AI}
This section describes a method for using generative AI to assist in the search for potential functions and the synthesis of mechanisms for the design problem of LA-GBMs.
Specifically, the general forms of the potential of LA-GBMs derived in the preceding sections are provided to the generative AI as design requirements, and candidate potential functions satisfying these requirements are generated.
The obtained candidates are then analytically verified to satisfy the design requirements. Furthermore, by associating the resulting functions with mechanical elements and linkage mechanisms that geometrically realize them, the possibility of mechanism synthesis using the potential field as an intermediate representation is investigated.

\subsection{Design Flow}
This section describes the design flow for LA-GBMs using generative AI.
In the proposed method, the affine form in Eq. (\ref{Eq_pr_4}), the factorized form in Eq. (\ref{Eq_new_1}), and the variation in the payload potential energy expressed by Eq. (\ref{Eq_pr_5}) are provided to the generative AI as design requirements representing the general forms and conditions that the potential field of an LA-GBM must satisfy.
The parameters required for the design are the dimensions $m$ and $n$ of the state variables $\bm{q}$ and $\bm{p}$, respectively.

Generative AI is used to search for candidate potential functions that satisfy these conditions, and the validity of the obtained functions is verified through the analysis described later.
Therefore, in the proposed method, generative AI is positioned not as the direct designer of the LA-GBM, but as a search tool for generating design candidates from a theoretically defined function space.

\begin{figure*}[t]
    \centering
    \includegraphics[width=0.7\linewidth, trim=0 140 0 140, clip]{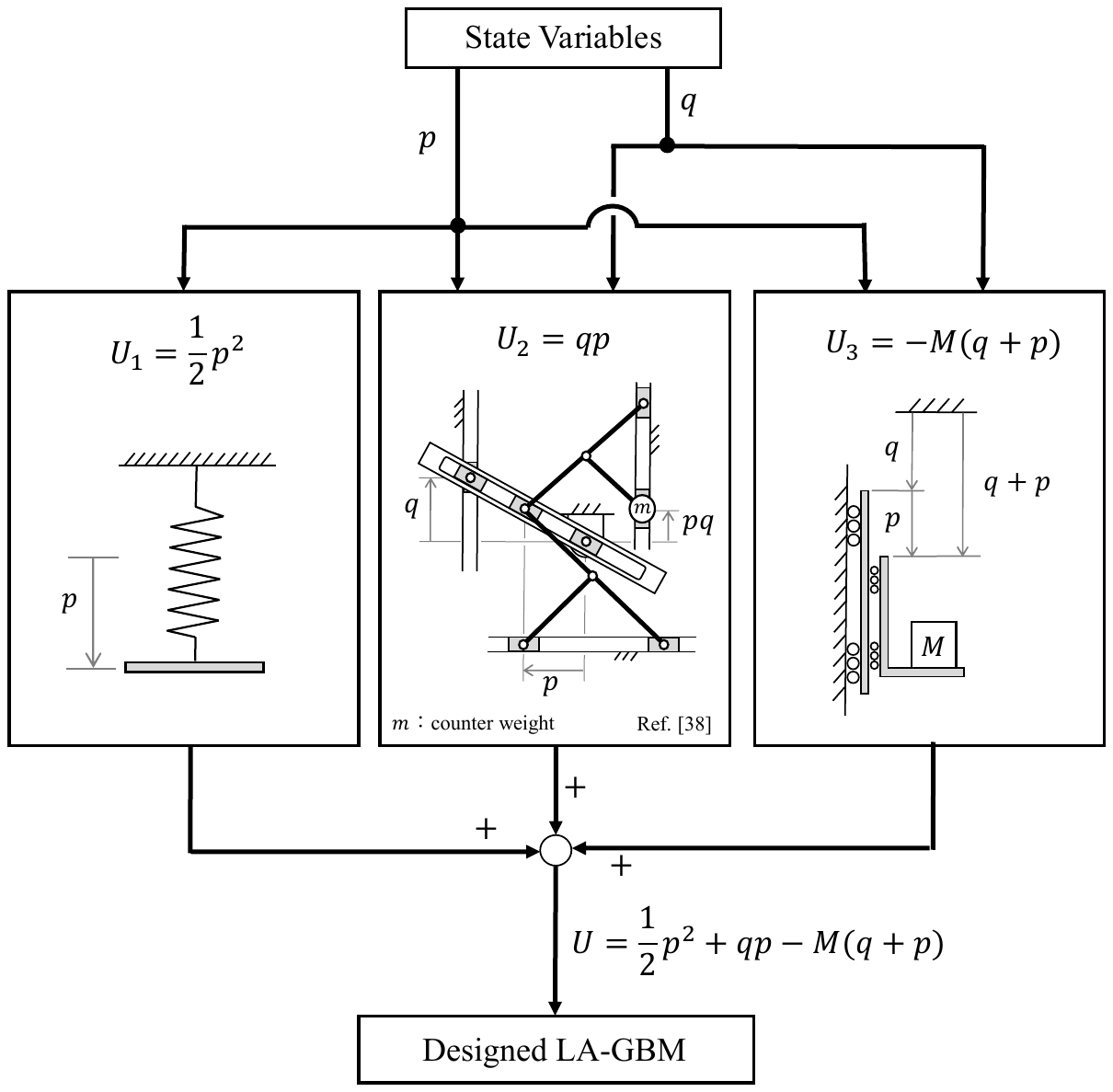}
    \caption{
    Design example of an LA-GBM based on the potential function obtained using generative AI into individual terms and combining the corresponding mechanical elements and linkage mechanisms in parallel.
    }
    \label{fig_design_flow}
\end{figure*}

\subsection{Example of Potential-Field Generation}
This section presents an example in which candidate potential fields for an LA-GBM were generated using generative AI.
For the generation of the potential field, GPT-5.6 Sol, a large language model provided by OpenAI, was used.

In this study, the model was used in a new conversation environment to minimize, as much as possible, the influence of prior conversation history and additional information specific to this study on the generated results.
No fine-tuning or additional training was performed for this study. Instead, the design requirements formulated in the preceding sections were provided as input conditions, and the model was asked to generate candidate potential functions satisfying these requirements. 
The generation was performed on September 17, 2026 (Japan Standard Time).
In addition, to obtain nontrivial solutions applicable to multi-degree-of-freedom systems, the active variable $\bm{q}$ was set as an $m$-dimensional real vector, $\bm{q}\in\mathbb{R}^m$, and the passive variable $\bm{p}$ as an $n$-dimensional real vector, $\bm{p}\in\mathbb{R}^n$.

The potential field $U$ generated by the AI is given as follows.
\begin{equation}
U(\bm{q},\bm{p};M)
=
\frac{1}{2}
\bm{p}^{\rm{T}}\bm{p}
+
\bm{q}^{\rm{T}} A \bm{p}
-
M \big(
\bm{b}^{\rm{T}} \bm{p}
+
\bm{q}^{\rm{T}} A \bm{b}
)
.
\label{Eq_AI_1}
\end{equation}
Here, $A \in \mathrm{R}^{m \times n}$ and $\bm{b} \in \mathrm{R}^{n}$ are an arbitrary constant matrix and an arbitrary constant vector, respectively, satisfying $A \bm{b} \neq \bm{0}$.
Figure \ref{fig_generated_motion} shows a schematic of the potential field in Eq. (\ref{Eq_AI_1}) for the case $\bm{q} \in \mathbb{R}$ and $\bm{p} \in \mathbb{R}$.

First, it is confirmed that Eq. (\ref{Eq_AI_1}) can be represented in both the affine form of Eq. (\ref{Eq_pr_4}) and the factorized form of Eq. (\ref{Eq_new_1}).
Equation (\ref{Eq_AI_1}) takes the affine form of Eq. (\ref{Eq_pr_4}) by defining the functions $f(\bm{q},\bm{p})$ and $h(\bm{q},\bm{p})$ as follows.
\begin{equation}
f(\bm{q},\bm{p})
=
-
\big(
\bm{b}^{\rm{T}} \bm{p}
+
\bm{q}^{\rm{T}} A \bm{b}
)
,
\label{Eq_AI_2}
\end{equation}
\begin{equation}
h(\bm{q},\bm{p})
=
\frac{1}{2} \bm{p}^{\rm{T}}\bm{p}
+
\bm{q}^{\rm{T}} A \bm{p}
.
\label{Eq_AI_3}
\end{equation}
Furthermore, when $\bm{t} \in \mathbb{R}^n$ is defined in the state $\bm{q} = \bm{0}$ as
\begin{equation}
\begin{aligned}
\bm{t}
&=
\nabla_{\bm{p}}U(\bm{0}, \bm{p};M)
=
\bm{p} - M \bm{b} \\
\therefore
\bm{p}
&=
\bm{t} + M \bm{b}
\end{aligned}
\label{Eq_AI_4}
\end{equation}
substituting Eq. (\ref{Eq_AI_4}) into Eq. (\ref{Eq_AI_1}) yields the following expression.
\begin{equation}
U(\bm{q},\bm{t};M)
=
\frac{1}{2}
\big(
\bm{t}^{\rm{T}}\bm{t}
-
M^2 \bm{b}^{\rm{T}}\bm{b}
\big)
+
\bm{q}^{\rm{T}} A \bm{t}
.
\label{Eq_AI_5}
\end{equation}
Equation (\ref{Eq_AI_5}) coincides with the factorized form expressed by Eq. (\ref{Eq_new_1}) by defining $\psi (\bm{t};M)$ and $B(\bm{q},\bm{t};M)$ as
\begin{equation}
\psi (\bm{t};M)
=
\frac{1}{2}
\big(
\bm{t}^{\rm{T}}\bm{t}
-
M^2 \bm{b}^{\rm{T}}\bm{b}
\big)
,
\label{Eq_AI_6}
\end{equation}
\begin{equation}
B(\bm{q},\bm{t};M)
=
A
.
\label{Eq_AI_7}
\end{equation}
It can also be confirmed that, when $\bm{t} = \nabla_{\bm{p}} U(\bm{0}, \bm{p};M) = \bm{0}$, Eq. (\ref{Eq_AI_5}) becomes
\begin{equation}
U(\bm{q},\bm{0};M)
=
-\frac{1}{2}M^2 \bm{b}^{\rm{T}}\bm{b}
=
\rm{const}
.
\label{Eq_AI_5_2}
\end{equation}
and therefore directly satisfies the condition in Eq. (\ref{Eq_pr_12}).

Furthermore, calculating the gradient $\nabla_{\bm{q}} f$ of Eq. (\ref{Eq_AI_2}) with respect to the active variable $\bm{q}$ gives
\begin{equation}
\nabla_{\bm{q}} f
=
-A \bm{b}
\neq
\bm{0}
.
\label{Eq_AI_8}
\end{equation}
which also satisfies the requirement for variation in the payload potential energy expressed by Eq. (\ref{Eq_pr_5}).

From the above results, it can be concluded that the function search using AI successfully generated a potential field satisfying the requirements of an LA-GBM.

\subsection{Example of Mechanism Design}
This section describes the process of constructing a gravity-balancing mechanism based on the potential field $U$ obtained in Sec. 3.2.
For simplicity, the dimensions of both the active variable $\bm{q}$ and the passive variable $\bm{p}$ are set to $1$, and they are denoted as the scalar quantities $\bm{q} = q$ and $\bm{p} = p$ for convenience.
Accordingly, the dimension of the variable $\bm{t}$ is also $1$, and it is denoted as $\bm{t} = t$.
For further simplification, $A = 1$ and $b = 1$ are assumed.
Under these assumptions, Eq. (\ref{Eq_AI_1}) is simplified as follows.
\begin{equation}
U(q,p;M)
=
\frac{1}{2} p^2
+
qp
-
M(q + p)
.
\label{Eq_AI_9}
\end{equation}
If the potential fields corresponding to the individual terms in Eq. (\ref{Eq_AI_9}) can be reproduced using existing mechanical elements, an LA-GBM can be designed based on the function generated by AI.
In the following discussion, all terms in Eq. (\ref{Eq_AI_9}) are assumed to have been converted to the dimension of energy using appropriate physical constants or conversion coefficients.

The potential field in Eq. (\ref{Eq_AI_9}) can be reproduced by combining a linear spring, a counterweight, and a linkage mechanism that outputs the product of two displacement inputs.
First, the first term on the right-hand side of Eq. (\ref{Eq_AI_9}) is a quadratic function of $p$ and can therefore be reproduced using a spring whose elongation is proportional to the state $p$.
Next, the second term on the right-hand side can be reproduced by constructing a linkage mechanism that generates a vertical displacement equal to the product $qp$ and attaching a counterweight to it. 
Several linkage mechanisms that output the product of state variables have previously been proposed \cite{seki_1,seki_2,seki_3}.
Finally, the third term on the right-hand side can be reproduced by constructing a linkage mechanism that generates a vertical displacement equal to the sum $q+p$ and attaching the payload $M$ to it.
By constructing a mechanism in which these three elements are connected in parallel, the potential field in Eq. (\ref{Eq_AI_9}) can be reproduced (Fig. \ref{fig_design_flow}).

Since the primary objective of this paper is to present a new design method for LA-GBMs, fabrication and experimental validation of the design result represented by Eq. (\ref{Eq_AI_9}) are left for future work.

\section{Discussion}
\begin{figure}[t]
    \centering
    \includegraphics[width=1.0\linewidth, trim=110 280 100 280, clip]{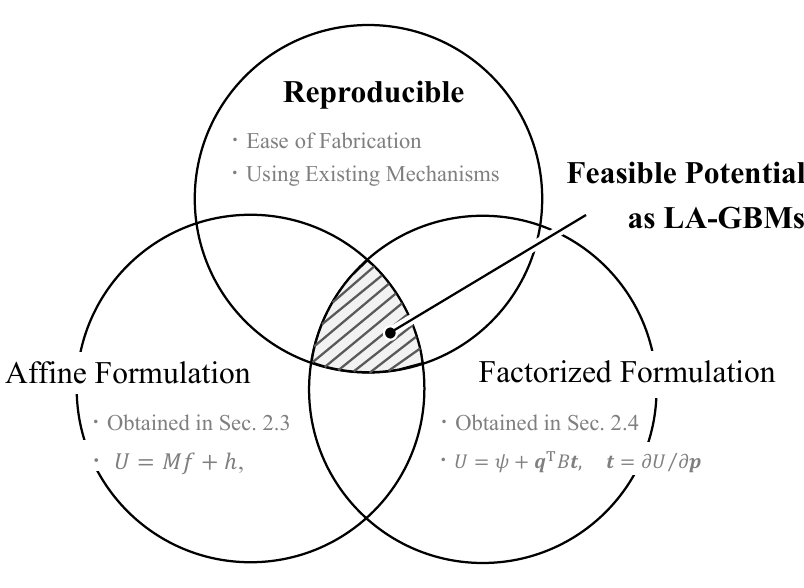}
    \caption{
    Set of LA-GBMs that can be constructed based on potential fields. Even among functions represented in the affine and factorized forms, some potential-field shapes may not be realizable using existing mechanical elements.
    }
    \label{fig_ben_diss}
\end{figure}

In this paper, the potential field formed by an LA-GBM was generalized in both affine and factorized forms, and a design method for LA-GBMs using potential functions as an intermediate representation was proposed by providing these conditions to generative AI as design requirements.
However, not all potential fields that can be generated mathematically by the proposed method can necessarily be reproduced easily using actual mechanical elements (Fig. \ref{fig_ben_diss}).
In other words, among the set of potential fields satisfying the affine and factorized forms, only a subset can be reproduced using existing mechanical elements such as linkage mechanisms, springs, and counterweights, and an even smaller subset is expected to satisfy practical implementation requirements such as a sufficiently large workspace and ease of fabrication.
For example, although the product $qp$ contained in the second term on the right-hand side of Eq. (\ref{Eq_AI_9}) can be reproduced by mechanisms that generate a displacement corresponding to the product of two state variables \cite{seki_1,seki_2,seki_3}, the resulting mechanism configuration and workspace are limited.

Various methods have previously been proposed for mechanically realizing functional relationships using function-generating mechanisms based on linkages and cams \cite{function_1,function_2}. 
Therefore, new LA-GBMs may be obtained by combining the potential functions generated in this study with existing function-generating mechanisms.
However, mathematical representability and the ability to construct a simple and practical mechanism are not necessarily equivalent, as illustrated by the example above.
This is an inherent challenge of the proposed method, in which a potential function is first generated and subsequently converted into a mechanism structure.
This issue may potentially be mitigated by incorporating existing techniques for mechanical function representation, as well as practical constraints such as workspace and manufacturability, into the design conditions provided to the generative AI.

\section{Conclusion}
In this study, the necessary conditions for the potential fields formed by many existing load-adaptive gravity-balancing mechanisms (LA-GBMs) with locking function were formulated and generalized using two approaches: the affine form and the factorized form. 
In addition, a potential-field-based design method for LA-GBMs was proposed by using generative AI to search for functions satisfying the derived affine and factorized forms as design requirements.

Using the proposed method, a potential field capable of realizing a nontrivial LA-GBM with $(m+n)$-dimensional state variables $(\bm{q},\bm{p})$ was generated. 
Furthermore, for the case in which the state variables are $\bm{q} \in \mathbb{R}$ and $\bm{p} \in \mathbb{R}$, the mechanism synthesis process for an LA-GBM based on the generated potential field was demonstrated through a specific example.

However, even when a potential field generated by AI mathematically satisfies the requirements for constructing an LA-GBM, it is not necessarily realizable as a simple mechanism.
Practical implementation constraints, such as the workspace and mechanism complexity, must also be considered.

Since this paper primarily focused on deriving general forms of the potential fields of LA-GBMs and demonstrating the feasibility of LA-GBM design by combining these formulations with AI, consideration of implementation constraints, fabrication of the generated mechanisms, and experimental evaluation of their performance are left for future work.



%
% ...

\bibliographystyle{asmems4}





\end{document}